\documentclass[10pt,journal,compsoc]{IEEEtran}
\ifCLASSOPTIONcompsoc
  \usepackage[nocompress]{cite}
\else
  \usepackage{cite}
\fi
\ifCLASSINFOpdf
\else
\fi
\usepackage{booktabs}
\usepackage{graphicx}
\usepackage{enumerate}
\usepackage{subfigure}
\usepackage{cite}

\usepackage{algorithm}
\usepackage{algorithmic}
\usepackage{graphics}

\usepackage{indentfirst}
\usepackage{verbatim}

\usepackage{amssymb}
\usepackage{longtable}
\usepackage{epsfig} 
\usepackage{amsthm}
\usepackage{mathrsfs}
\usepackage{amsmath}
\usepackage{stfloats}
\usepackage{multirow}
\usepackage{caption}
\usepackage[hang,flushmargin]{footmisc}
\newtheorem{theorem}{Theorem}

\begin{document}
%
\title{ADDS: Adaptive Differentiable Sampling for Robust Multi-Party Learning}
%
%
%
%

\author{Maoguo~Gong,~\IEEEmembership{Senior Member,~IEEE,}
  Yuan~Gao,
  Yue~Wu,
  and~A.~K.~Qin,~\IEEEmembership{Senior Member,~IEEE}\\

}

%
%

\markboth{}%
{Shell \MakeLowercase{\textit{et al.}}: Bare Demo of IEEEtran.cls for Computer Society Journals}
%



\IEEEtitleabstractindextext{%
\begin{abstract}
Distributed multi-party learning provides an effective approach for training a joint model with scattered data under legal and practical constraints. However, due to the quagmire of a skewed distribution of data labels across participants and the computation bottleneck of local devices, how to build smaller customized models for clients in various scenarios while providing updates appliable to the central model remains a challenge. In this paper, we propose a novel adaptive differentiable sampling framework (ADDS) for robust and communication-efficient multi-party learning. Inspired by the idea of dropout in neural networks, we introduce a network sampling strategy in the multi-party setting, which distributes different subnets of the central model to clients for updating, and the differentiable sampling rates allow each client to extract optimal local architecture from the supernet according to its private data distribution. The approach requires minimal modifications to the existing multi-party learning structure, and it is capable of integrating local updates of all subnets into the supernet, improving the robustness of the central model. The proposed framework significantly reduces local computation and communication costs while speeding up the central model convergence, as we demonstrated through experiments on real-world datasets.

\end{abstract}

\begin{IEEEkeywords}
multi-party learning, distributed learning, model customization, adaptive sampling.
\end{IEEEkeywords}}

\maketitle

\IEEEdisplaynontitleabstractindextext

%
\IEEEpeerreviewmaketitle

\IEEEraisesectionheading{\section{Introduction}\label{para:1}}

%
%
%
%

\IEEEPARstart{W}{ith} data increasingly being generated on various remote devices, there is a growing interest in building models that make use of such data. However, it becomes unrealistic to integrate the scattered data due to the enhancement of people's awareness of privacy and the improvement of relevant data policy restrictions, making it hard to realize effective artificial intelligence applications. To alleviate this challenge, a novel learning paradigm \cite{Bonawitz2019TowardsFL} \cite{Yang2019FederatedML} that empowers knowledge sharing among multiple parties is presented, whose main idea is to aggregate models from participants to achieve an efficient central model, and it enables collaborations among clients with private non-shared data.

In the past few years, distributed multi-party learning has gained growing popularity \cite{Sun2020LazilyAQ,Lu2020BlockchainAF,Xu2020VerifyNetSA,Kang2020ReliableFL}. A preliminary investigation is given in \cite{Chen2016RevisitingDS}. They proposed a synchronous approach named Federated Stochastic Gradient Descent (FedSGD), where each client conducts a single batch gradient computation per round of communication and sends updated gradients to the server for aggregation. Build upon the technique of FedSGD, McMahan \emph{et al.} \cite{McMahan2017CommunicationEfficientLO} introduced Federated Averaging (FedAvg) that allows more local training iterations before the averaging step. Since clients provide higher-quality model parameters compared with FedSGD, FedAvg speeds up the convergence of the central model, and thus becomes the typical representative of multi-party learning algorithms.

Most existing approaches are confronted with a contradiction between the heterogeneity of individual private data and the generality of the central model. Specifically, as users are with different lifestyles and environments, clients' local data is usually distributed in a non-IID manner \cite{Sattler2020RobustAC}, so that a large and complex central model is required to learn the distinct distributions and summary the local knowledge. However, this deep neural network presents a dilemma, which is that the model can be hardly adapted to a small amount of biased and specific local data on each client without further fine-tuning \cite{Jiang2019ImprovingFL,Li2019FedMDHF}, and it is along with high computation and communication costs. Even worse, global collaboration without considering the characteristics of individual private data usually cannot accomplish good performance for individual clients \cite{Huang2020PersonalizedCF}, greatly reducing the enthusiasm of clients to participate in the collaboration. Unfortunately, the problem is widespread in scenarios where the server and clients deploy the same network architecture. Some local customization methods, e.g. knowledge distillation \cite{Lin2020EnsembleDF,Wang2021KnowledgeDA} and multi-task learning \cite{Smith2017FederatedML,Sattler2020ClusteredFL}, provide personalized models with different structures for each client to adapt the individual private data, whereas the obstacle of how to aggregate models effectively remains unresolved.

To address the aforementioned challenges, the paper explores the dropout strategy in a multi-party setting and proposes an effective adaptive differentiable sampling framework, named ADDS, for robust and communication-efficient multi-party learning, as illustrated in Fig.~\ref{fig:frame}. As its name suggests, ADDS implies that the global model is a dynamic integration of local models. Following the underlying idea of global collaboration, an over-parameterized and dense supernet is deployed on the server. Participants measure the importance of hidden units (neurons or channels) in the supernet according to their private data distribution, and sample individually customized subnets from the supernet. Moreover, we investigate an adaptive sampling scheme by introducing differentiable and trainable sampling rates, which ensure each participant derives the optimal model architecture. The local optimizations for model parameters and structure are performed alternatively, and the subnet with recorded index map is merged into the supernet. The network sampling reduces complicated co-adaptive relationships among hidden units in the supernet and hence improves the robustness. Meanwhile, it lowers the computation and communication costs for clients without degrading the network quality. Specifically, the main contributions of this paper are as follows:
\begin{figure*}[htb!]
  \centering
  \includegraphics[width=0.88\linewidth]{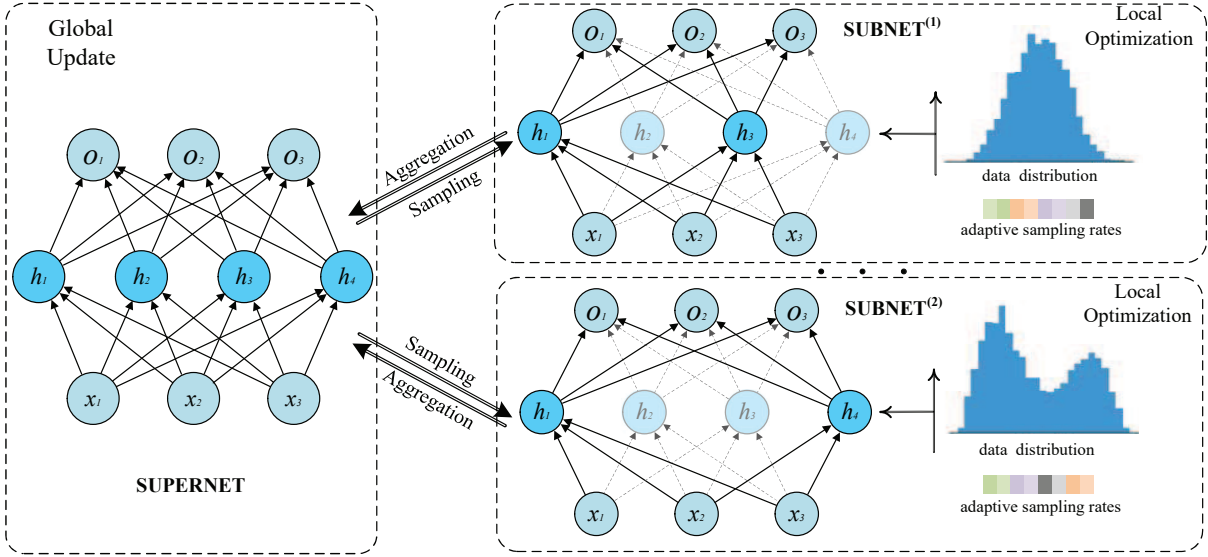}
  \caption{
  Diagram of the adaptive differentiable sampling framework in a multi-party setting.}
\label{fig:frame}
\end{figure*}
\begin{itemize}
\item{The paper introduces a network sampling scheme to provide solutions for multi-party learning problems, and we show that it is a natural choice for handling the challenge of statistical heterogeneity and model customization in the multi-party setting.}
\item{The paper provides a novel approach for deriving a personalized model for each client, and the embedded adaptive differentiable sampling rates empower the gradient descent optimization for both model parameters and network architecture.}
\item{The supernet-subnet design makes it possible to build a customized model dynamically while providing updates applicable to the central model, which is capable of improving the robustness and generalization ability of the dense network while reducing the local computation and communication costs simultaneously.}
\item{The proposed ADDS framework is extensible to various effective deep neural networks, since it only requires minimal modifications to the existing multi-party learning scheme. Experimental results on real-world image and text datasets demonstrate that our framework has superior performance over the state-of-the-art baselines.}
\end{itemize}

The remainder of this paper is organized as follows. Section~\ref{para:2} briefly presents the related backgrounds about multi-party learning, several personalization techniques and structured network pruning. In Section~\ref{para:3}, we give a clear definition of the multi-party learning problem, then the details of our framework are described. Section~\ref{para:4} shows extensive experiments to validate the effectiveness. Finally, we conclude with a discussion of our framework and summarize the future work in Section~\ref{para:5}.

\section{Background and Related Works}
\label{para:2}
\subsection{Multi-Party Learning}
\label{para:2.1}
The number of research in multi-party learning, which facilitates collaborations among clients and learn a global model while keeping private data locally, is proliferating in recent years. It takes the form that all participants download the latest central model from the server and train it on their local private data, then upload the updated models to the server, where these models are gathered and aggregated to form a new central model. The interactions are repeated until the central model reaches the convergence criterion, and the local training data never leaves clients during the communications, and hence the risk of privacy leakage is minimized.

A typical implementation of this work is the Federated Averaging (FedAvg), which iterates the local updates $w^k_{t+1}=w^k_t-\eta g_k$ multiple times for each client $k$, where $g_k=\nabla F_k(w_t)$ represents the gradient, and takes a weighted average of the model parameters on the server:
\begin{equation}
w_{t+1} = \sum\nolimits^K_{k=1}\frac{n_k}{n}w^k_{t+1}
\end{equation}
Furthermore, it is observed by McMahan \emph{et al.} \cite{McMahan2017CommunicationEfficientLO} that the approach is surprisingly well-behaved when deploying sufficiently over-parameterized neural networks in each client, and a wide enough model can be less prone to bad local minima.

However, though over-parameterized models achieve high performance with scattered individual data that follows the same distribution \cite{AllenZhu2019LearningAG}, it faces the obstacles of statistical heterogeneity in the multi-party setting, i.e. the unbalanced and non-IID data, which can lead to serious performance degradation \cite{Li2020OnTC}. The skewed distribution of data labels across participants violates the underlying assumption of deep learning that stochastic gradient is an unbiased estimate of the full gradient, resulting in a significant reduction in the testing accuracy and thus limits the applications of multi-party learning. 

To approach the non-IID challenge, Zhao \emph{et al.} \cite{Zhao2018FederatedLW} investigated the weight divergence due to the skewness of the data distribution and found that the test accuracy falls sharply as the Wasserstein distance between the distributions increases. Thus, they proposed to distribute a globally shared dataset containing a uniform distribution over classes to clients to reduce the Wasserstein distance. Chiu \emph{et al.} \cite{Chiu2020SemisupervisedDL} also introduced a federated swapping operation that allows clients to exchange local models based on a few shared data during training. However, it is mentioned in \cite{Sattler2020RobustAC} that overfitting to the shared data is a serious issue if all clients share the same public dataset. 

\subsection{Personalization Techniques}
Since the non-IID data greatly decreases the aggregation efficiency in multi-party learning, it removes the incentive for clients to participate as they gain no benefit. That is, though prior works focused on measuring the overall accuracy of the aggregated model, the global accuracy cannot represent the performance of the model on a specific client when participants own their individual idiosyncratic data, and the central model can be worse than local for the majority of clients \cite{Yu2020SalvagingFL}. To tackle this problem, there is a growing trend \cite{Dinh2020PersonalizedFL,Hu2020PersonalizedFL,Huang2020PersonalizedCF,Deng2020AdaptivePF} to convert the central model into customized models for each client by local adaptation, instead of building a single model that is supposed to be well-behaved for all participants. There are many personalization techniques for adapting the central model to an individual participant, and several main directions are listed as follows:

\noindent\emph{\textbf{Local fine-tuning.}} Fine-tuning is predominantly used in domain adaption \cite{Song2020RetrainingSD}, transfer learning \cite{Luo2019TransferringKF} and Model Agnostic Meta Learning (MAML) \cite{Wang2020OnTG}, where the knowledge acquired before is transferred into a relevant subspace. It is a natural adaption and dominant approach for personalization in multi-party learning, which re-trains parameters of the central model on individual private data. A typical variant is a freeze-based scheme, which freezes the base layers of the central model and fine-tunes the top layers to take advantage of the feature extraction network. Jiang \emph{et al.} \cite{Jiang2019ImprovingFL} analyzed the similarity between the setting of MAML and the objective of local fine-tuning and presented multi-party learning as a natural source of practical applications for MAML algorithms. Another approach is introduced by Khodak \emph{et al.} \cite{Khodak2019AdaptiveGM}, who employed online convex optimization to instantiate a meta learning approach in multi-party learning for better personalization. The main drawback of local fine-tuning is that the customized model is prone to overfit individual data since it minimizes the local optimization error and limits the generalization.

\noindent\emph{\textbf{Multi-task learning.}} Another significant trail for personalization is to view the multi-party learning as a multi-task learning problem \cite{Smith2017FederatedML,Song2020AnalyzingUP}, where the optimization on each party is a subtask. In view of the massive number of participants and the difficulty of joint optimization, there are other studies to cluster groups of clients based on uploaded model parameters or some features as similar tasks \cite{Sattler2020ClusteredFL,Huang2020PersonalizedCF,Zhou2021APD}. While the multi-task learning scheme provides a natural choice to handle statistical challenges in multi-party learning by learning separate models, jointly solving the weight and relationships among tasks can be difficult and the paradigm is hard to apply to non-convex deep learning models \cite{Smith2017FederatedML}.

\noindent\emph{\textbf{Knowledge distillation.}} Knowledge distillation \cite{Park2019RelationalKD,Han2020NeuralCM} aims to transfer knowledge acquired in a teacher model to a student model. In the multi-party setting, the central model is treated as the teacher while customized models as students \cite{Lin2020EnsembleDF,Wang2021KnowledgeDA}. A small dataset is deployed on the server to assist the knowledge distillation, and the method applies the regularization and enforces the similarity of logits on the predictions between the two models. The extra benefit is that it alleviates the heterogeneity of both data and models among clients in multi-party learning. However, the performance is limited due to the defects of knowledge distillation itself \cite{Wang2021KnowledgeDA} as well as the overfitting with the common dataset.

\subsection{Network Pruning}
Communication is a critical bottleneck in a distributed multi-party learning framework, which is compromised of a massive number of clients, and the large-capacity communication among clients and the server even drags down the computational efficiency. To reduce the communication in such a setting, in addition to speed up the convergence of the central model, model compression at each round is necessary as well. Sattler \emph{et al.} \cite{Sattler2020RobustAC} proposed sparse ternary compression with top-$n$ gradient sparsification for both downstream and upstream communication. The Count Sketch is introduced by Rothchild \emph{et al.} \cite{Rothchild2020FetchSGDCF} to compress model updates, and they combined model updates from clients leveraging the mergeability of sketches. The prior works focused on reducing the communication costs through naive lossy compression, whereas do not consider common characteristics of the multi-party learning, e.g. locally-updating optimization approaches or low device participation \cite{Li2020FederatedLC}.

Network pruning by removing the redundant connections provides a valid way to reduce the network complexity and overfitting \cite{LeCun1989OptimalBD,Liu2019RethinkingTV,He2020AsymptoticSF,Ning2020DSAME}. Especially, structured network pruning, which prunes at the levels of neurons and convolution channels, improves the efficiency of over-parameterized neural networks for applications with a limited computational budget while not significantly degrading the model performance. It is usually conducted on trained deep networks to find the most informative hidden units for the predictions, and the other connections are safely dropped according to a certain criterion. Fine-tuning is also required in most existing techniques to retrain the model to regain the lost accuracy.

Structured pruning tallies naturally with characteristics of multi-party learning since it reduces the expense of local computation and communication. Besides, it could effectively avoid overfitting on unbalanced and non-IID data through sampling different subnet at each round. However, as a hyperparameter, the fixed pruning rate limits the potential of network pruning \cite{Caldas2018ExpandingTR}, and the local data for participants is private and cannot be revealed as prior knowledge to estimate the parameter. Therefore, we introduce an adaptive sampling scheme with trainable pruning rates (or sparseness) for local model customization, so that each client can derive a personalized network on individual data. To the best of our knowledge, this paper is the first to explore the differentiable architecture adaption and joint optimization in a multi-party setting.

\section{Proposed Algorithm}
\label{para:3}
\subsection{Problem Statement}
Consider there are $N$ clients, each holding a small set of private data that can only be stored and processed locally. The goal of our multi-party learning framework is to build a central statistical model from scatted data across participants, as well as customized models adapted for each client. In this setting, the central model minimizes the empirical risk on the aggregated domain

\begin{equation}
\underset{{w\in \mathbb{R}^d}}{\rm min}\ f(w)\quad {\rm where}\quad f(w)\overset{\rm def}=\frac{1}{N}\sum\limits_{i=1}^N f_i(\theta).
\end{equation}

The function $f_i:\mathbb{R}^d\rightarrow\mathbb{R},\ i=1,...,N$, reprsents the expected loss over the distribution of client $i$, and $\theta_i$ is the corresponding customized model. It typically takes
\begin{equation}
f_i(\theta)=\mathbb{E}_{\xi_i}[\widetilde f_i(\theta;\xi_i)],
\end{equation}
where $\xi_i$ denotes a random sample drawn from the distribution of client $i$, and $\widetilde f_i(\theta;\xi_i)$ is the loss function for the sample and $\theta_i$. To achieve the goal of improving computation and communication efficiency, a simplified but adapted local model $\theta_i$ is supposed to be derived for each participant $i$. Meanwhile, an effective indexed aggregation approach for different model structures should be explored to avoid the degradation of the performance of the central model $w$.

\subsection{Bilevel Optimization}
In multi-party learning, building customized models with different architectures enables participants to be well-behaved on unbalanced and non-IID data, whereas it raises the difficulty to aggregate these models together effectively. The regularisation strategy of dropout \cite{Srivastava2014DropoutAS} provides a natural thought for model sampling and aggregation in the multi-party setting. Specifically, an over-parameterized supernet similar to the conventional works is deployed on the server, but each participant trains an adapted sparse subnet sampled from the supernet instead of training the whole model, and the updates are filled back into the supernet by associated index map. 

\begin{algorithm}[!t]
  \caption{ ADDS - Adapted Differentiable Sampling}
  \begin{algorithmic}[1]
  \label{alg:server}
  \REQUIRE
  communication rounds; number of local iterations; $N$ clients indexed by $i$
  \ENSURE
  global supernet $w$, local subnets $\theta$
  \\ \hspace*{\fill} \\
  \textbf{Server executes:}
  \STATE Initialize the over-parameterized supernet $w$;\\
  \FOR{each communication round}
     \STATE Send supernet $w$ to each client $i$;
     \FOR {each client $i$ \textbf{in parallel}}
      \STATE Receive subnet $\theta_i$ and index map $\mathit{I}_i$ from local clients;
      \ENDFOR
     \STATE Update $w$ with all $\theta$ and associated $\mathit{I}$;
  \ENDFOR
  \RETURN global supernet $w$
  \\ \hspace*{\fill} \\
  \textbf{Client executes in each round:}
  \STATE Receive supernet $w$ from the server;
  \STATE Initialize the local architecture $\mathcal{A}$;
  \FOR{each local iteration}
    \STATE Update $\mathcal{A}$ by descending $\nabla_\mathcal{A}\mathcal{L}_{val}(w^*(\mathcal{A}),\mathcal{A})$
    \STATE Update weights $\theta$ by descending $\nabla_w\mathcal{L}_{train}(w,\mathcal{A})$
  \ENDFOR
  \STATE Derive the final architecture based on the learned $\mathcal{A}$, send the associated weights $\theta$ and index map $\mathit{I}$ to the server;
  \RETURN local subnets $\theta$
  \end{algorithmic}
\end{algorithm}

The derivation of the subnet can be easily achieved through structured pruning, where the most informative hidden units for the predictions are kept while the others are safely dropped according to a certain criterion, and a sparse network is thus built. However, the obstacle lies in the determination of the sparseness. Since the skewed data distribution among participants, a fixed sparseness for all clients requires detailed prior knowledge of individual privacy data, which is impractical due to regulatory restrictions, and the sensitive hyperparameter may affect the performance for both local and central models. 

Inspired by DARTS \cite{Liu2019DARTSDA}, which conducts network architecture search (NAS) by jointly optimizing both weights and differentiable architecture, ADDS regards the local model customization as an optimization problem using gradient descent. The individual private data is split into training and validation datasets, where the losses are denoted by $\mathcal{L}_{train}$ and $\mathcal{L}_{val}$ respectively, both of which are determined by the architecture and weights of the subnet. Suppose the sparseness $\mathcal{A}=\{\alpha^k\}_{k=1,...,K}$ is the keep ratio of $K$ layers that ranges from 0 to 1, indicating the architecture of the subnet. Without losing generality, denote the task loss as $\mathcal{F}(w,\mathcal{A})$, and the losses can be written as:
\begin{align}
&\mathcal{L}_{train}=\mathcal{F}_{train}(w,\mathcal{A}) \\
&\mathcal{L}_{val}=\mathcal{F}_{val}(w,\mathcal{A})+\lambda\mathcal{R}(\mathcal{A}),
\end{align}
where $\mathcal{R}(\mathcal{A})$ is the regularization term on sparseness, which is introduced in a simple form:
\begin{equation}
\mathcal{R}(\mathcal{A})=\sum\nolimits_{k=1}^K||\alpha^k||_2^2.
\end{equation}
It achieves a balance between the performance and architecture of the subnet. $\lambda\geq0$ is a client-specific hyperparameter and a larger $\lambda$ leads to a smaller architecture. The goal of the local optimization of ADDS is to find $\mathcal{A}^*$ that minimizes the $\mathcal{L}_{val}(w^*(\mathcal{A}),\mathcal{A}^*)$, where the corresponding weights $w^*(\mathcal{A})$ are optimized by minimizing $w^*={\rm argmin}_w\ \mathcal{L}_{train}(w,\mathcal{A}^*)$, and we have $\theta=w^*(\mathcal{A})$ for the subnet. The joint optimization problem is fomulated as:
\begin{align}
&\min\limits_\mathcal{A}\quad \mathcal{L}_{val}(w^*(\mathcal{A}),\mathcal{A}) \\
&{\rm s.t.} \quad w^*(\mathcal{A})={\rm argmin}_w\ \mathcal{L}_{train}(w,\mathcal{A}),
\end{align}
which implies a bilevel optimization problem with $\mathcal{A}$ as the upper-level variable and $w$ as the lower-level variable, and the optimizations for model parameters and architecture constitute Stackelberg Game \cite{Fiez2019ConvergenceOL}, where the architecture serves as the leader while parameters as the follower. Two optimization objectives achieve local optimum alternatively and assist the update of each other, in order to find the subgame perfect Nash equilibrium, i.e. a well-trained subnet with adapted architecture. The workflow of ADDS is depicted in Alg.~\ref{alg:server}. 

Due to the expensive inner optimization when evaluating $w^*(\mathcal{A})$, a first-order approximation scheme in \cite{Liu2019DARTSDA} is adopted to adapt the weights to the changes of the architecture by ignoring the high order gradients $\nabla_\mathcal{A}w^*(\mathcal{A})$. That is,
\begin{equation}
\nabla_\mathcal{A}\mathcal{L}_{val}(w^*(\mathcal{A}),\mathcal{A})\approx \nabla_\mathcal{A}\mathcal{L}_{val}(w,\mathcal{A}),
\end{equation}
where $w$ denotes the current weights, and the architecture gradient equals the simple heuristic of optimizing the $\mathcal{L}_{val}$ by assuming the current $w$ is the same as $w^*(\mathcal{A})$. It is a reasonable assumption in the multi-party setting as the subnet converges fast on the small amount of local data, and thus $w$ is near to the local optimum for the inner optimization.

\subsection{Soft Relaxtion and Differentiable Sampling}
\label{sec:33}
Analogous to conventional network pruning that a subset of hidden units is dropped with a given ratio $\alpha$ according to the importance criteria $b_c\in \mathbb{R}^+,c=1,...,C$, the sampling of hidden units in ADDS is associated with the importance as well. Layer-wise Relevance Propagation (LRP) \cite{Yeom2021PruningBE} is leveraged to assign relevance scores to channels and neurons in neural networks. The relevance is backpropagated from the output to the input, which essentially reflects the importance of every single hidden unit and its contribution to the information flow through the network. Let $R_i^l$ be the relevance of neuron $i$ at layer $l$, and the layer-wise conservation principle satisfies
\begin{equation}
\sum\nolimits_i R^l_{i \gets j} = R_j^{l+1},
\end{equation}
which allows the propagated quantity to be preserved between two adjacent layers during inference. The propagation rule can be written as
\begin{equation}
R_i^l=\sum\nolimits_j\frac{(a_i^l w_{ij})^+}{\sum\nolimits_{i'}(a_{i'}^l w_{i'j})^+}R_j^{l+1},
\end{equation}
where $i'$ runs over all activations $a$. The equation indicates an aggregation of backpropagated relevance messages $R^{(l,l+1)}_{i\gets j}$, which is locally conservative. Therefore, the relevance can be interpreted as the contribution each unit makes to the output, and thus units supporting the model decision for classes of interest could be preserved.

A significant difference between conventional network pruning and network sampling in ADDS lies in the probabilistic soft relaxation employed in our scheme, where each hidden unit $c$ is kept with a probability $p_c$ to facilitate the derivation, so that the sparseness for each layer is trainable and the architecture could be more flexible. For the following steps, we take channels in convolutional layers as an example, and all deductions can be adapted to neurons.

Given an initialized sparseness set $\mathcal{A}=\{\alpha^k\}_{k=1,...,K}$, the channel-wise sampling rate $p_c$ is associated with its improtance $b_c$ and the sparseness $\alpha$ of this layer, i.e. we have $p_c=f(\alpha,b_c)$. Then, a subnet is sampled from the supernet with each $p_c$, and the sampling results follow the Bernoulli distribution: $\omega_c \sim {\rm Bernoulli}(p_c)$. It should be expected that
\begin{equation}
\alpha C=E[\sum\nolimits^C_{c=1}\omega_c]=\sum\nolimits^C_{c=1}p_c=\sum\nolimits^C_{c=1}f(\alpha,b_c)
\label{eq:ca}
\end{equation}
which indicates the proportion of sampled channels equals the sparseness $\alpha$. Since the importance $b_c\in \mathbb{R}^+$ while $\alpha\in(0,1]$, a shift factor $\beta$ that maps $b_c$ to an appropriate range $b_c-\beta$ is required to guarantee the sparseness condition. Besides, to derive the final discrete subnet architecture, we introduce a hyperparameter $\varepsilon$ to control the inexactness of the sampling process, hence the top-$n$ informative channels with the highest importance $b_c$ can be retained eventually. To summarize, a function $f(b_c,\beta,\varepsilon):\mathbb{R}^+\to(0,1)$ with two parameters is required, and thus we can control the sampling exactness and sparseness simultaneously.

\begin{figure}[tb!]
  \centering
  \includegraphics[width=1.0\linewidth]{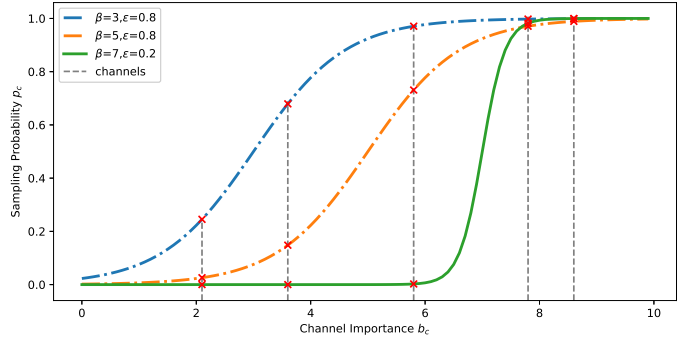}
  \caption{
  Illustration of the effects of $\beta$ and $\varepsilon$ on the sampling probability. The three colored lines are derived by different sampling rates and sparseness $\beta(\alpha)$, where the blue and orange dotted lines indicate the soft sampling with a larger $\varepsilon$, and the green solid line represents the hard sampling with a smaller $\varepsilon$.}
\label{fig:logsigmoid}
\end{figure}
The logistic sigmoid function is commonly used to produce the $\phi$ parameter of a Bernoulli distribution, and it lies within the valid range of values for the $\phi$ parameter. The sigmoid function saturates when its argument is very large or small, controlled by the shift factor $\beta$, facilitating the hard sampling of informative or useless channels. To achieve the desired property, the function is defined as
\begin{equation}
f(b_c,\beta,\varepsilon)=\frac{1}{1+e^{-\frac{b_c-\beta}{\varepsilon}}},\quad \beta,\varepsilon>0,
\end{equation}
in which $\beta$ ensures the sparseness and $\varepsilon$ controls the inexactness. To be specific, $\beta=\beta(\alpha)$ is a relative function of $\alpha$, which determines the proportion of sampled channels through by shifiting the importance $b_c$. The value of $\beta$ can be calculated by solving the implicit equation of the sparseness guarantee:
\begin{equation}
h(\beta)=E[\sum\nolimits^C_{c=1}\omega_c]-\alpha C=\sum\nolimits^C_{c=1}f(b_c,\beta,\varepsilon)-\alpha C=0.
\end{equation}
The function $h(\beta)$ is monotonically decreasing because the probability $p_c$ for each channel $c$ decreases with the increase of $\beta$, as illustrated in Fig.~\ref{fig:logsigmoid}. Therefore, the root $\beta(\alpha)$ can be simply found in several efficient ways, e.g. the binary search algorithm. Moreover, the other parameter, the inexactness $\varepsilon$, follows a decreasing schedule in practice. It is related to the slope of the $S$-curve, which is sensitive to $\beta(\alpha)$ when $\varepsilon$ approaches 0. In this way, the sampling degrades to hard network pruning and the top-$n$ informative channels with the highest importance $b_c$ can be retained eventually.

\begin{figure*}[htb!]
  \centering
  \includegraphics[width=0.96\linewidth]{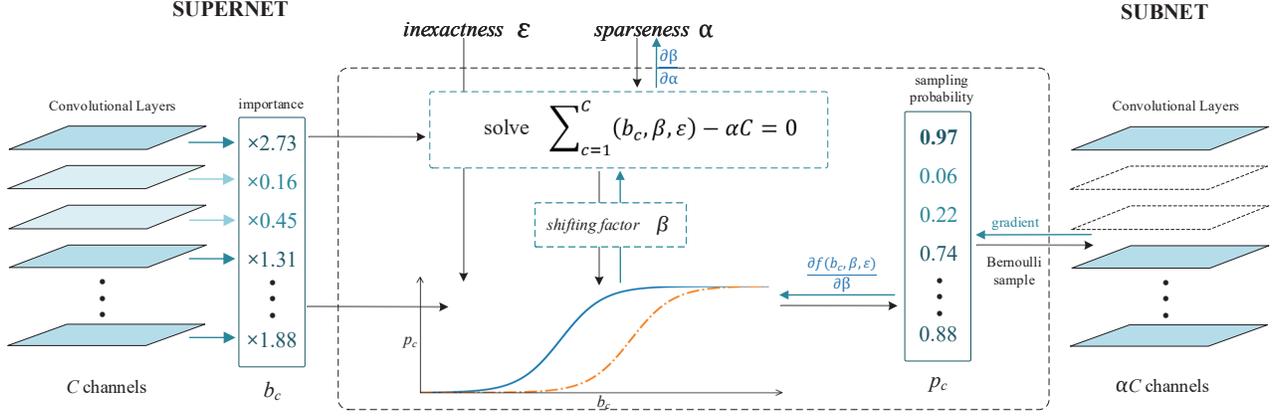}
  \caption{
  Workflow of the network sampling strategy and the gradient backward scheme, where inexactness $\varepsilon$ is a hyperparameter that follows a decreasing schedule, sparseness $\alpha$ is a trainable variable, and the shifting factor $\beta$ is an intermediate result associated with $\alpha$.}
\label{fig:prune}
\end{figure*}

To solve the partial derivative of the local loss with respect to $\alpha$, i.e., $\frac{\partial \mathcal{L}_{val}}{\partial\alpha}$ ($k$ is moitted for simplicity), the implicit gradient $\frac{\partial \beta(\alpha)}{\partial\alpha}$ is supposed to be calculated firstly according to Eq.~\ref{eq:ca}:
\begin{align}
&\sum\nolimits^C_{c=1}\frac{\partial f(b_c,\beta,\varepsilon)}{\partial\beta} \cdot \frac{\partial\beta}{\partial\alpha}=C,\\
&\frac{\partial\beta(\alpha)}{\partial\alpha}=\frac{C}{\sum\nolimits^C_{c=1}\nabla_\beta f(b_c,\beta,\varepsilon)},
\end{align}
in which we have
\begin{equation}
\frac{\partial f(b_c,\beta,\varepsilon)}{\partial\beta}=\nabla_\beta f(b_c,\beta,\varepsilon)=\frac{p_c(1-p_c)}{\varepsilon}.
\end{equation}
Thus, the partial derivative can be derived as:
\begin{equation}
\begin{split}
\frac{\partial \mathcal{L}_{val}}{\partial\alpha}&=\sum\nolimits^C_{c=1}\frac{\partial \mathcal{L}_{val}}{\partial p_c} \cdot \frac{\partial p_c}{\partial\beta} \cdot \frac{\partial\beta}{\partial\alpha}\\
&=C\sum\nolimits^C_{c=1}\frac{\partial \mathcal{L}_{val}}{\partial p_c} \cdot \frac{p_c(1-p_c)}{\sum\nolimits^C_{c=1} p_c(1-p_c)},
\end{split}
\end{equation}
and the shifting factor $\beta(\alpha)$ is recalculated as well along with the updates of sparseness $\alpha$. The computation of the partial derivative $\frac{\partial \mathcal{L}_{val}}{\partial p_c}$ may raises some concerns due to the Bernoulli sampling results $\omega$. To solve this problem we use straight-through estimators in the backward pass, and it ignores the derivative of the threshold function and passes on the incoming gradient, whose idea is naturally adapted for the sparse subnets. Furthermore, note that for the Bernoulli distribution, we have $\sigma^2(\omega_c)=p_c(1-p_c)$, and that means the updates of $\alpha$ is associated with the sampling probability $p_c$ of channels and the corresponding variances $\sigma^2(\omega_c), c=1,...,C^k$. 

The network sampling strategy and the gradient backward scheme when optimizing $\mathcal{L}_{val}$ are visualized in Fig.~\ref{fig:prune}. A customized subnet is sampled from the supernet for each client, and the gradient backpropagation in a single iteration is conducted on the sampled parameters conveniently by introducing a mask, so that the subnet is updated through iterative sampling flow without the requirement of pre-training.

\subsection{Indexed Aggregation and Analysis}
The adaptive sampling strategy is equivalent to identifying the winning lottery tickets \cite{Frankle2019TheLT} in the supernet, which makes the training procedure particularly effective, and ADDS framework can be characterized as simultaneously training the ensemble of all winning tickets (subnets) dependent on the local private samples with different individual data distributions.

The bilevel optimization enables each client to derive adapted architecture while updating local weights. Obtaining the well-trained subnet $\theta$, the client is supposed to record the corresponding index map $\mathit{I}$ of it in the supernet, then uploads $\theta$ and $\mathit{I}$ to the server. The index map $\mathit{I}$ can be instantiated as a tensor that contains the indexes of sampled hidden units in a derived subnet. To be specific, a tensor of length $J_w$ that full of ones is initialized as the index map $\mathit{I}_w=[1,1,...,1]$ of the supernet $w$ with $J_w$ representing the number of hidden units. Note that it indicates the units that can be sampled and is different from the number of parameters, where the former is far less than the latter. The same operation is conducted for a subnet $\theta$. When recording the index map $\mathit{I}_\theta=[1,0,...,1]$ of a subnet $\theta$ with $J_\theta$ ($J_w>J_\theta$) hidden units, the length of tensor $\mathit{I}_\theta$ is $J_w$ as well, whereas there are $J_\theta$ ones in the tensor and the others are zeros.

Thus, a simple but effective indexed aggregation approach is leveraged to update the supernet, which is formulated as follow:
\begin{equation}
w_j^{t+1}=\left\{
\begin{aligned}
&\frac{\sum\nolimits_{i=1}^N I_{\theta i}[j]\cdot\theta^t_j}{\sum\nolimits_{i=1}^N I_{\theta i}[j]}\quad {\rm if}\ \ \exists\ I_{\theta i}[j]\ \ \ {\rm s.t.}\ I_{\theta i}[j]=1 \\
\\
& w_j^t \quad {\rm otherwise}
\end{aligned}
\right.
\label{equ:update}
\end{equation}
indicating the averaging aggregation of each hidden units $w_j$ in $t+1$ communication round. The operation can be easily implemented by matrix multiplication for both channels and neurons without gaining computation complexity in the central server. Besides, the size of index map $\mathit{I}$ is positively correlated with the number of channels and neurons in the neural network, usually within the range of 1e+3$\sim$1e+4 bits. In view of the extremely large number of parameters of the model (1e+7$\sim$1e+8 floating points), that of $\mathit{I}$ can be negligible compared with the communication costs saved by the sampled subnets.

In addition to the size of networks to be transmitted between the clients and server, local FLOPs required for the convergence of the customized model is a key factor affecting the computation costs in multi-party learning as well. We first investigate the convergence of the local bilevel optimization, which can be guaranteed in Theorem~1.
\begin{theorem}
Let $\max \mathcal{F}(w,\mathcal{A})<\infty$ and $\mathcal{F}$ be differentiable, then the sequence of $\mathcal{A}$ generated along with local iterations has limit points.
\end{theorem}
\noindent\emph{Proof.} While the boundedness assumption is made, the smoothness of $\mathcal{F}$ can be satisfied using proper loss functions, such as the Cross-Entropy Loss in the experimental section. The task loss $\mathcal{F}_{val}$ (and validation loss $\mathcal{L}_{val}$) is proved to be differentiable with respect to $\mathcal{A}$ in Section~\ref{sec:33}. Hence, by the definition of $\mathcal{F}$, the loss is continuous on $\mathcal{A}$ and bounded from below. Given $\mathcal{A}=\{\alpha^k|0<\alpha^k\leq1\}$, the sequence of $\mathcal{A}$ is constrained with a compact sublevel set, and thus the Theorem comes from the fact that any infinite sequence on a compact sub-level set has at least one limit point.

In summary, due to the limited capacity and small individual dataset of the remote devices, training an over-parameterized model on the clients could be less effective. Local personalization will improve the situation and yield superior performance for each participant, yet the knowledge aggregation among various customized models becomes a major obstacle. The proposed ADDS adopts a supernet-subnet design and introduces an adaptive differentiable sampling to allow participants to sample adapted subnets from the supernet according to their private non-IID data distribution, and it enables the efficient bilevel optimization of model parameters and architecture using gradient descent. ADDS is capable of deriving the optimal subnet for each client with increased accuracy, and the local model with fewer Floating Point of Operations (FLOPs) significantly facilitates the subsequent applications. Besides, the effective indexed aggregation drawing on the idea of dropout improves the generalization of the central model, and thus speeds up the convergence as well as reduces the communication cost.

\begin{table*}[tbp]
\small
\renewcommand{\arraystretch}{1.3}
  \centering
  \caption{Statistics of selected datasets.}
  \setlength{\tabcolsep}{4mm}{
    \begin{tabular}{cccccccc}
    \cline{1-8}
      \multirow{2}{*}{\bf{Dataset}}&\multirow{2}{*}{\bf{Clients}}&\multirow{2}{*}{\bf{Samples}}&\multirow{2}{*}{\bf{Classes}}&\multicolumn{2}{c}{\bf{samples per client}}&\multicolumn{2}{c}{\bf{classes per client}}\\
      \cline{5-8}
      &&&&mean&sedev&min&max\\
      \cline{1-8}
      FEMNIST&1,062&241,756&62&286&78&9&62\\
      FedCelebA&1,213&47,903&2&29&11&1&2\\
      Shakespeare&528&625,127&70&1183&1218&2&70\\
      Sentiment140&1,330&57,385&2&43&19&1&2\\
      \cline{1-8}
    \end{tabular}}
\label{tab:data}
\end{table*}
\section{Experiments}
\label{para:4}
\subsection{Experimental Settings}
\subsubsection{Datasets}
The experiments are conducted on four open-source LEAF \cite{Caldas2018LEAFAB} datasets for practical multi-party learning environments, so that the performance of the proposed ADDS framework can be evaluated with various tasks and models. The statistics of the selected datasets are summarized in Table~\ref{tab:data}.

\noindent\textbf{FEMNIST} for 62-class image classification. It is built by partitioning the digit or character in the Extended MNIST dataset \cite{Cohen2017EMNISTAE} based on the writer. 

\noindent\textbf{FedCelebA} for 2-class image recognition. It partitions the Large-scale CelebFaces Attributes Datasets \cite{liu2015faceattributes} by the celebrity on images. We extract face images for each person and target to recognize whether the person smiles or not.

\noindent\textbf{Shakespeare} for next word prediction. It is built from The Complete Works of William Shakespeare \cite{McMahan2017CommunicationEfficientLO}, and each speaking role in each play is considered as a different client.

\noindent\textbf{Sentiment140} for 2-class sentiment classification. It annotates tweets based on the emoticons present in them, where each client represents a different Twitter user.

\subsubsection{Baselines}
The experiments investigate the testing accuracy on both global and individual datasets, convergence speed and computation cost, and these measures are derived from different methods for comparison. The centralized learning with mini-batch gradient descent serves as the baseline, which gathers all data in the server and enables the central model to train on it without privacy concerns. The unbalanced and non-IID conditions are greatly alleviated, so that the baseline usually yields superior performance. Besides, the conventional multi-party learning algorithm, i.e. the Federated Averaging (FedAvg) \cite{McMahan2017CommunicationEfficientLO}, is compared with the proposed framework as well, which can be regarded as a degradation form of ADDS that fixes each $\alpha^k$ to 1. The ADDS generates a customized subnet for each client, whereas all local models for FedAvg are identical. 

In view of the personalized and communication-efficient characteristics of ADDS, we compare the comprehensive performance with other state-of-the-art algorithms. STC \cite{Sattler2020RobustAC} is a compression framework, named sparse ternary compression, which enables downstream compression as well as ternarization and optimal Golomb encoding of the weight updates. PFedMe \cite{Dinh2020PersonalizedFL} uses Moreau envelopes as clients' regularized loss functions and helps decouple personalized model optimization from the global model. The computation complexity $K$ is set 5, which allows for approximately finding the personalized model. FedDrop \cite{Caldas2018ExpandingTR} drops a fixed percentage of neurons randomly and clients train the same smaller model in each round. The drop rate is fixed at 0.75, which works across the board in their experiments. To measure the effectiveness of ADDS on individual data, local models of all approaches are evaluated before the parameter aggregation in each communication round.

\begin{table*}[tbp]
\small
\renewcommand{\arraystretch}{1.8}
  \centering
  \caption{Model architectures of CNNs.}
  \setlength{\tabcolsep}{4mm}{
    \begin{tabular}{ccc}
      \hline
      Architecture&VGG&MobileNet\\
      \hline
      Convolutional&64, pool, 128, pool, 256, pool&64, pool, 128, pool, 256, pool, 512, pool, 512, pool\\
      \hline
      Fully-connected&1024, 1024, 62 (input size: 4096)&1024, 1024, 2 (input size: 8192)\\
      \hline
      Conv/FC/all params&369.22K/5.31M/5.68M&444.63K/9.50M/9.95M\\
      \hline
      Conv/FC/all FLOPs&29.37M/5.31M/34.68M&114.96M/9.50M/124.46M\\
      \hline
    \end{tabular}}
\label{tab:arch}
\end{table*}
\subsubsection{Models}
Various neural networks are deployed for different datasets to validate the performance of the proposed ADDS. For the FEMNIST dataset, we follow the architecture of the VGG-like network and build a supernet with three $3\times 3$ convolutional layers and three fully-connected layers, in which the ReLU-Conv-BN order is utilized for convolutional operations. Each of the convolutional layers is followed with $2\times 2$ max pooling. The model architecture is listed in Table~\ref{tab:arch}, which shows the number of parameters and FLOPs of convolutional layers and fully-connected layers as well.

For the FedCelebA dataset, we employ an efficient MobileNet \cite{Howard2017MobileNetsEC} for mobile vision applications to simulate the multi-party learning scenario. Compared with the standard convolutions in VGG-like models, MobileNet uses depthwise separable convolutions to build light weight networks. To be specific, it factorizes a standard convolution into a depthwise convolution and a $1\times1$ pointwise convolution, and applies a single filter to each input channel. A significant reduction in computation is achieved by expressing convolution as a two steps process, and we will investigate how the ADDS works on it. The supernet consists of five $3\times 3$ convolutional layers, each followed with $2\times 2$ max pooling, and three fully-connected layers. The convolutional operation is the same as we described above for FEMNIST.

The sampling strategy is appliable to the non-recurrent layers for RNNs so that the memorization ability could be preserved. Thus, the architecture of RNNs for text processing is not listed in the Table. For the Shakespeare dataset, the supernet maps each word to an embedding of 32 dimensions, and passes it through a two-layer bi-directional LSTM of 256 hidden units. A sequence length of 80 is set for LSTM, and its output embedding is scored against a vocabulary followed by a softmax. A similar model is used for the Sentiment140 dataset, while the main difference is that pre-trained 300-dimensional GloVe embeddings \cite{Pennington2014GloveGV} are leveraged. A sequence length of 25 is set for LSTM, and the output of the last fully-connected layer is a binary classifier.

\subsubsection{Detailed Setup}
In all experiments, 80\% of clients are used for training and 20\% for evaluating the performance of the central model. For individual data in training clients, we also set 80\% of data as training set and the remaining as the test set. Besides, 10\% of the training data is split for the validation set in ADDS to update $\mathcal{A}$ and 90\% to tune the weights. Note that samples in the validation set are not fixed for a client. Besides, the total number of communication rounds for all datasets is set at 200. We randomly select 30\% of the clients for training in each round, and the number of local epochs is set to be 3 for all methods.

For the adaptive differentiable sampling, the $L_1$ norm of the zoom factor $\gamma$ in batch normalization layer and activations in fully-connected layer serve as the importance $b$ of channels and neurons, respectively. Two key impacting hyperparameters are inexactness $\varepsilon$ and the client-specific weight scalar $\lambda$. The former is associated with the sampling probability and follows a decreasing schedule, e.g., starts at 1 and gets multiplied by 0.98 every round. The latter affects the architecture of subnets, and we have $\lambda={\rm Norm}(JSD)+0.5$, whose value is related to the normalized Jensen-Shannon divergence \cite{Jin2018PredictingAS} (JSD) between the skewed distribution of local labels and the discrete uniform distribution, assuming gathered data is balanced and IID.

In order to evaluate the effectiveness of ADDS when handling the challenge of statistical heterogeneity, extensive experiments are conducted on real-world datasets. We investigate the communication and computation costs, then conduct systems analysis. In addition, the correlation between data distribution and model architecture is demonstrated as well. In the experiments, we repeat each experiment for 5 trials then report the averages and variances.

\subsection{Accuracy Comparison}

\begin{figure*}[htb!]
  \centering
  \includegraphics[width=0.98\linewidth]{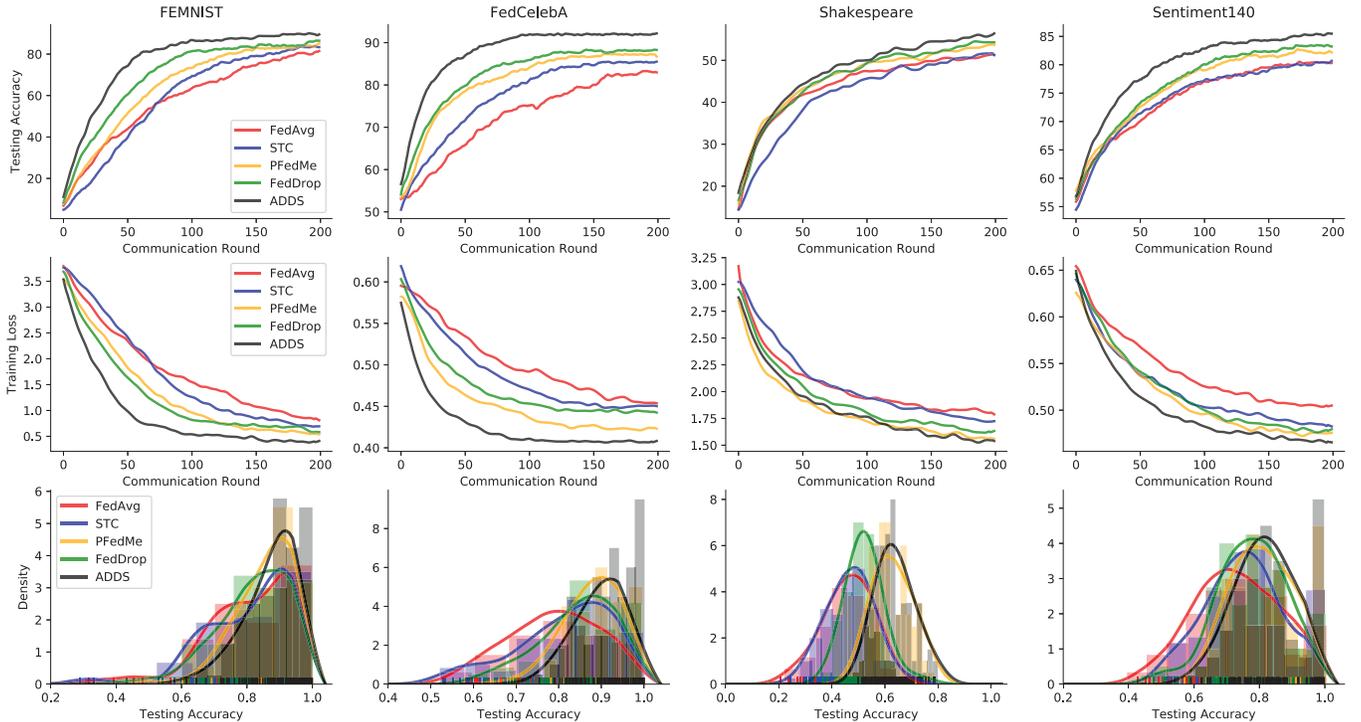}
  \caption{
  The global and local accuracy on four datasets. Compared with intuitive FedAvg and three state-of-the-art algorithms, ADDS framework provides faster convergence and higher accuracy.}
\label{fig:all}
\end{figure*}

In order to highlight the empirical performance of ADDS, we perform several comparisons between ADDS, FedAvg, STC, PFedMe and FedDrop. The model performance of different multi-party learning algorithms in testing accuracy, training loss and density is evaluated at first, as illustrated in Fig.~\ref{fig:all}, to demonstrate the superiority of ADDS for both global and local models.

In terms of testing accuracy, STC makes a slight improvement over FedAvg, whereas its convergence rate slows down when facing multi-classification datasets, such as FEMNIST and Shakespeare. PFedMe and FedDrop are well-performed than FedAvg as well. The former makes use of the Moreau envelope regularized loss function and obtains the convergence speedup, while the latter enables each device to locally operate on the same smaller model and improve the model performance with the idea of dropout. It is observed that the proposed ADDS framework outperforms FedAvg and other state-of-the-art algorithms in both accuracy and convergence rate on all datasets, where the improvement is more significant on image datasets, since all units in CNNs can be sampled and the subnet architectures are more flexible. Each client trains on its customized subnet instead of a random one while providing available updates to the supernet, combining the strengths of both personalized techniques and dropout strategy.

The training loss, which is obtained by averaging all clients' losses, is roughly negatively correlated with the testing accuracy. In particular, by decoupling the process of optimizing personalized models from learning the global model through the Moreau envelope function, PFedMe achieves a decrease in training loss, which indicates that the local performance gains cannot be merged into the global model. ADDS optimizes the weights and architecture of local models with respect to each client's individual data distribution, and the smaller architecture further contributes to better speedup convergence rates. 

We also investigate the accuracy of the local test set and depict the density in Fig.~\ref{fig:all}, where the vertical lines represent the accuracy of each client and different algorithms are distinguished by color. The bar chart indicates the number of clients within a certain accuracy range, the width of which is associated with the variance of the approach. The density curve is smoothed for a better visual effect. The local accuracy for personalized multi-party learning algorithms, i.e., PFedMe and ADDS, fluctuates less compared with methods that distribute the global model to all participants. In particular, the proposed ADDS framework performs slightly better with higher overall local accuracy, because each subnet converges faster with smaller customized network architecture, demonstrating its advantages as achieving the highest global and local accuracy, as well as the smallest training loss.

\subsection{Comprehensive Performance Analysis}
\begin{table*}[tbp]
\small
\renewcommand{\arraystretch}{1.6}
  \centering
  \caption{Performance and model statistics of different multi-party learning algorithms on image datasets.}
  \setlength{\tabcolsep}{5mm}{
    \begin{tabular}{c|cccc}
    \multicolumn{5}{c}{\textbf{FEMNIST}}\\ \hline
      \bf{Method}&Acc$_G$&Acc$_L$&Parameters&FLOPs\\ \hline
      FedAvg     &81.36$\%\ \pm$ 0.29$\%$&84.64$\%\ \pm$ 0.31$\%$&5.68M          &34.68M\\ \hline
      STC        &83.65$\%\ \pm$ 0.53$\%$&86.25$\%\ \pm$ 0.41$\%$&1.39M (24.39\%)&34.61M (1 $\times$)\\ \hline
      PFedMe     &83.22$\%\ \pm$ 0.78$\%$&92.23$\%\ \pm$ 0.55$\%$&5.68M (100.0\%)&34.61M (1 $\times$)\\ \hline
      FedDrop    &84.97$\%\ \pm$ 0.43$\%$&87.56$\%\ \pm$ 0.68$\%$&2.62M (46.07\%)&19.05M (1.82 $\times$)\\ \hline\hline
      ADDS (slim)&88.95$\%\ \pm$ 0.84$\%$&91.21$\%\ \pm$ 0.75$\%$&0.99M (17.45\%)&16.95M (2.04 $\times$)\\ \hline
      ADDS (LRP) &\bf{90.25$\%\ \pm$ 0.93}$\%$&\bf{93.12$\%\ \pm$ 0.69}$\%$&\bf{0.85M (15.06\%)}&\bf{16.13M (2.15 $\times$)}\\ \hline
    \multicolumn{5}{c}{\textbf{FedCelebA}}\\ \hline
      \bf{Method}&Acc$_G$&Acc$_L$&Parameters&FLOPs\\ \hline
      FedAvg     &83.41$\%\ \pm$ 0.11$\%$&90.75$\%\ \pm$ 0.75$\%$&9.95M          &124.46M\\ \hline
      STC        &85.86$\%\ \pm$ 0.56$\%$&91.62$\%\ \pm$ 0.26$\%$&1.57M (15.82\%)&124.46M (1 $\times$)\\ \hline
      PFedMe     &87.29$\%\ \pm$ 0.84$\%$&95.46$\%\ \pm$ 0.96$\%$&9.95M (100.0\%)&124.46M (1 $\times$)\\ \hline
      FedDrop    &88.61$\%\ \pm$ 0.47$\%$&93.48$\%\ \pm$ 0.46$\%$&4.41M (44.36\%)&84.67M (1.47 $\times$)\\ \hline\hline
      ADDS (slim)&91.98$\%\ \pm$ 1.06$\%$&96.10$\%\ \pm$ 0.82$\%$&1.37M (13.78\%)&57.48M (2.17 $\times$)\\ \hline
      ADDS (LRP) &\bf{92.41$\%\ \pm$ 0.92}$\%$&\bf{96.90$\%\ \pm$ 0.71}$\%$&\bf{1.31M (13.16\%)}&\bf{56.20M (2.21 $\times$)}\\ \hline
    \end{tabular}}
\label{tab:img}
\end{table*}

\begin{table*}[tbp]
\small
\renewcommand{\arraystretch}{1.6}
  \centering
  \caption{Performance and convergence rate of different multi-party learning algorithms on text datasets.}
  \setlength{\tabcolsep}{3.6mm}{
    \begin{tabular}{c|cc|ccc}
    \multicolumn{6}{c}{\textbf{Shakespeare}}\\ \hline
      \multirow{2}{*}{\bf{Method}}&\multirow{2}{*}{Acc$_G$}&\multirow{2}{*}{Acc$_L$}&\multicolumn{3}{c}{\textbf{Convergence Rounds to Target Accuracy}}\\ \cline{4-6}
      &&&40\%&45\%&50\%\\ \hline
      FedAvg     &51.89$\%\ \pm$ 0.68$\%$&53.97$\%\ \pm$ 0.74$\%$      &41                &78                &155\\ \hline
      STC        &52.71$\%\ \pm$ 1.22$\%$&54.63$\%\ \pm$ 0.72$\%$      &54 (0.76 $\times$)&94 (0.83 $\times$)&166 (0.93 $\times$)\\ \hline
      PFedMe     &53.10$\%\ \pm$ 0.83$\%$&60.02$\%\ \pm$ 0.95$\%$      &38 (1.08 $\times$)&64 (1.22 $\times$)&121 (1.28 $\times$)\\ \hline
      FedDrop    &54.48$\%\ \pm$ 0.87$\%$&57.24$\%\ \pm$ 0.78$\%$      &40 (1.03 $\times$)&63 (1.24 $\times$)&107 (1.45 $\times$)\\ \hline\hline
      ADDS (slim)&\bf{56.13$\%\ \pm$ 0.71}$\%$&59.51$\%\ \pm$ 0.81$\%$ &\bf{34 (1.21 $\times$)}&\bf{53 (1.47 $\times$)}&\bf{98 (1.58 $\times$)}\\ \hline
      ADDS (LRP) &55.97$\%\ \pm$ 0.62$\%$&\bf{60.16$\%\ \pm$ 0.89}$\%$ &\bf{34 (1.21 $\times$)}&55 (1.42 $\times$)&101 (1.53 $\times$)\\ \hline
    \multicolumn{6}{c}{\textbf{Sentiment140}}\\ \hline
      \multirow{2}{*}{\bf{Method}}&\multirow{2}{*}{Acc$_G$}&\multirow{2}{*}{Acc$_L$}&\multicolumn{3}{c}{\textbf{Convergence Rounds to Target Accuracy}}\\ \cline{4-6}
      &&&70\%&75\%&80\%\\ \hline
      FedAvg     &80.46$\%\ \pm$ 0.12$\%$&84.62$\%\ \pm$ 0.39$\%$&50          &83&159\\ \hline
      STC        &80.98$\%\ \pm$ 0.58$\%$&84.19$\%\ \pm$ 0.31$\%$&42 (1.19 $\times$)&76 (1.09 $\times$)&170 (0.93 $\times$)\\ \hline
      PFedMe     &82.29$\%\ \pm$ 0.41$\%$&90.15$\%\ \pm$ 0.80$\%$&40 (1.25 $\times$)&65 (1.28 $\times$)&111 (1.43 $\times$)\\ \hline
      FedDrop    &83.01$\%\ \pm$ 0.56$\%$&87.36$\%\ \pm$ 0.52$\%$&37 (1.35 $\times$)&62 (1.34 $\times$)&99 (1.61 $\times$)\\ \hline\hline
      ADDS (slim)&84.96$\%\ \pm$ 0.63$\%$&89.63$\%\ \pm$ 0.69$\%$&24 (2.08 $\times$)&41 (2.02 $\times$)&72 (2.21 $\times$)\\ \hline
      ADDS (LRP) &\bf{85.13$\%\ \pm$ 0.71}$\%$&\bf{90.37$\%\ \pm$ 0.74}$\%$&\bf{22 (2.27 $\times$)}&\bf{37 (2.24 $\times$)}&\bf{66 (2.41 $\times$)}\\ \hline
    \end{tabular}}
\label{tab:txt}
\end{table*}

The comprehensive performance analysis including more intuitive global and local accuracy (denoted as Acc$_G$ and Acc$_L$ respectively) is conducted to validate the effectiveness of ADDS. Besides the aforementioned LRP strategy, a more simple but popular method, network slimming \cite{Liu2017LearningEC}, is leveraged to measure the importance of channels and neurons. To be specific, it introduces a batch normalization layer after each convolutional layer and regards the associated zoom factor $\gamma$ in batch normalization layer as the channel importance, and the activation in fully-connected layer serves as the neuron importance. 

Table~\ref{tab:img} shows the number of parameters and FLOPs of CNNs on image datasets. Compared with the intuitive FedAvg algorithm, STC reduces both upstream and downstream communication necessary which achieving higher accuracy. PFedMe achieves significant improvement on local accuracy with multiple local gradient updates, which promotes global accuracy as well. ADDS distinctively outperforms other state-of-the-art approaches on the four measurements. By customized network sampling for each participant, ADDS is more efficient and effective than FedDrop, which also adopts the dropout strategy. In addition, LRP works better than network slimming as the relevance scores reflect the information flow through the network, and it yields superior performance on the standard convolutional model and depth-wise separable convolutional model.

Since the sampling strategy is merely applied to the non-recurrent layers for RNNs to preserve the memorization ability, the compression capability for the customized models is not obvious like that on CNNs, and hence we investigate the convergence rounds to the target accuracy, as listed in Table~\ref{tab:txt}. Network slimming yields competitive performance with LRP in this scenario, though it is less well-performed than PFedMe in terms of local accuracy. ADDS converges more stable and faster than other methods, and it even achieves more than twice the acceleration ratio compared with FedAvg on sentiment analysis.

ADDS are more advantageous in situations where the communication is bandwidth-constrained or costly. It promotes clients with limited communication and computing conditions to participant in collaborative multi-party learning by significantly reducing local model parameters and FLOPs, which also facilitates the deployment of customized models on remote devices. Besides, the sparse subnets uploaded by each client and the partial parameter aggregation scheme on the server could prevent model inversion attacks from adversary semi-honest participants \cite{Zhang2020TheSR}, so that it further preserves individual privacy. The characteristic of ADDS is out of the scope of this paper and will be investigated in our future work.

\subsection{Adaptive Sampling}

\begin{figure*}[htb!]
  \centering
  \includegraphics[width=0.88\linewidth]{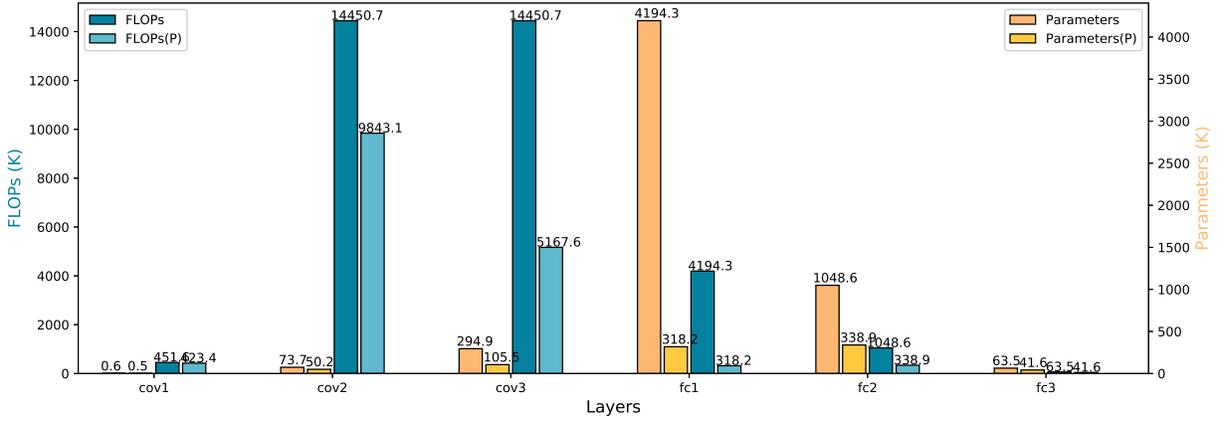}
  \caption{
  Effect of layer-wise adaptive sampling rate on the quantity of parameters and FLOPs.}
\label{fig:prune}
\end{figure*}

\begin{figure}[tb!]
  \centering
  \includegraphics[width=1.0\linewidth]{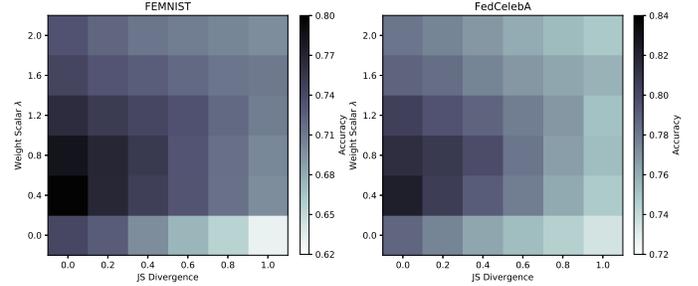}
  \caption{
  Accuracy achieved in a $distributed$ setting at different levels of normalized JS divergence and fixed weight scalar $\lambda$. In each column, the JS divergence between label distribution for all clients (around 2\% training clients are selected) is the same, and the performance of the supernet with different $\lambda$ is evaluated.}
\label{fig:heat1}
\end{figure}

The proposed network sampling strategy works at both layer and model levels. At the layer level, it provides flexible and adaptive sampling rates for each layer of CNNs. Take the model for FEMNIST as an example, Fig.~\ref{fig:prune} depicts the effect of layer-wise adaptive sampling rate on the number of parameters and FLOPs. The shallow layers near the input layer are responsible for extracting low-order information, which is available for data of different classes, so that discarding these channels may lead to severe performance degradation. The network sampling is mainly concentrated in the deeper network layer accompanied by more channels and neurons, which extract high-order features and combine them for specific classes of data. Due to the skewed distribution of labels in the non-IID setting, most units are futile and merely result in the increase of model size. Therefore, through the differentiable sampling strategy, the ADDS framework is capable of adopting lower sampling rates for layers with more units, pursuing the minimum model size without hurting the testing accuracy.
\begin{figure}[tb!]
  \centering
  \includegraphics[width=1.0\linewidth]{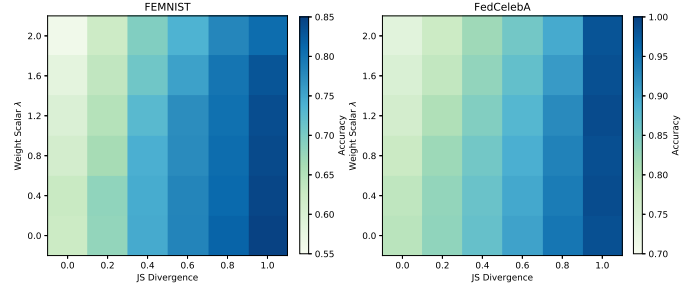}
  \caption{
  Accuracy achieved in a $centralized$ setting at different levels of normalized JS divergence and fixed weight scalar $\lambda$. In each column, a client with specific JS divergence between label distribution is randomly selected, and its performance without parameter aggregation is evaluated as well.}
\label{fig:heat2}
\end{figure}

The discussion of adaptive sampling can be more in-depth at the model level. Considering different levels of normalized JS divergence between the skewed distribution of local labels and the discrete uniform distribution of gathered data, the accuracy achieved in a distributed and centralized setting with fixed weight scalar $\lambda$ is illustrated in Fig.~\ref{fig:heat1} and Fig.~\ref{fig:heat2}, respectively. For both scenarios and datasets, the testing accuracy decreases with the growth up of the JS divergence, whereas the situation changes with different weight scalar. In the distributed setting, when $\lambda$ equals 0, ADDS degrades to the conventional FedAvg algorithm and the performance is poor. A larger $\lambda$ causes fewer network units to be sampled, which hurts the accuracy when local data tends to be independent and identically distributed. However, it surprisingly increases the accuracy with larger JS divergence (0.8 in this case, where $\lambda$=1.2 yields superior performance), which reaches a delicate balance between model size and performance. In the centralized setting, i.e., deep learning on a single client without collaboration, network sampling results in performance degradation to varying degrees, and the effect will weaken as the JS divergence increases. Hence, ADDS discards the fixed setting of $\lambda$ and keeps the value relevant to the local label distribution, so as to obtain optimal performance.
\begin{figure}[tb!]
  \centering
  \includegraphics[width=1.0\linewidth]{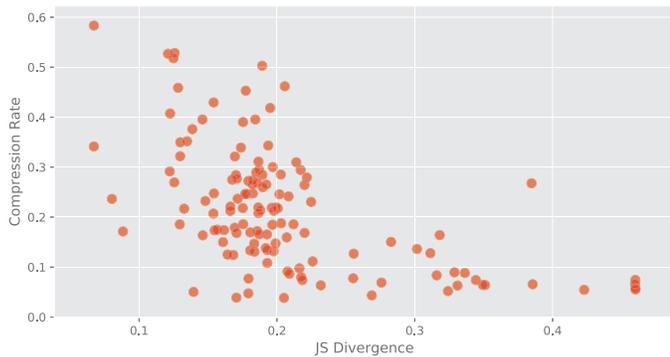}
  \caption{
  The compression rates of customized models and the associated JS Divergence on the FEMNIST dataset.}
\label{fig:scatter}
\end{figure}

The compression rates of individual customized models and the associated JS Divergence on the FEMNIST dataset are shown in Fig.~\ref{fig:scatter}. It is observed that there is a negative correlation between compression rates and the JS divergence in general, though other factors, such as the quality of local data, affect the compression rate as well. The statistic empirically validates the effectiveness of the proposed adaptive sampling strategy on the macro level, and the model customization can be achieved by the distribution-oriented sampling.

\section{Conclusion}
\label{para:5}
In this work, we develop a novel adaptive differentiable sampling framework (ADDS) for robust and communication-efficient multi-party learning. To address the problem of statistical and model personalization in a conventional multi-party setting, we introduce the network sampling scheme to derive a customized model for each client. Moreover, the supernet-subnet design enables efficient parameter aggregation while reducing local communication and computation costs. Extensive experimental results on several real-world datasets with different neural networks demonstrate its effectiveness in various meaningful tasks.

We will explore the following directions in the future:

(1) We have validated the effectiveness of ADDS in convolutional layers and fully-connected layers, and we will make it scalable for recurrent layers, as well as investigate a more effective sampling strategy, to improve the scalability of ADDS.

(2) The proposed ADDS achieves superior performance in a variety of scenarios and significantly reduces communication costs. However, the bilevel optimization increases the computational complexity in the training process, and the supernet is required to be distributed in each round, which heavier burdens of clients with limited resources. We try to explore some new approaches for leveraging computational resources in the server to further facilitate the model training and deployment in remote devices.


%





\ifCLASSOPTIONcaptionsoff
  \newpage
\fi


\bibliographystyle{IEEEtran}
\bibliography{ref}
\end{document}